# Retrieving Similar X-Ray Images from Big Image Data Using Radon Barcodes with Single Projections

Morteza Babaie[1,2], H.R. Tizhoosh[1], Shujin Zhu[3], and M.E. Shiri[2]

[1]*KIMIA Lab, University of Waterloo, ON, CANADA*
[2]*SINA Lab, Mathematics & Computer Science Department, Amirkabir University of Technology, Tehran, IRAN*
[3]*School of Electronic & Optical Eng., Nanjing Univ. of Sci. & Tech., Jiangsu, CHINA*

Abstract: The idea of Radon barcodes (RBC) has been introduced recently. In this paper, we propose a content-based image retrieval approach for big datasets based on Radon barcodes. Our method (Single Projection Radon Barcode, or SP-RBC) uses only a few Radon single projections for each image as global features that can serve as a basis for weak learners. This is our most important contribution in this work, which improves the results of the RBC considerably. As a matter of fact, only one projection of an image, as short as a single SURF feature vector, can already achieve acceptable results. Nevertheless, using multiple projections in a long vector will not deliver anticipated improvements. To exploit the information inherent in each projection, our method uses the outcome of each projection separately and then applies more precise local search on the small subset of retrieved images. We have tested our method using IRMA 2009 dataset a with 14,400 x-ray images as part of imageCLEF initiative. Our approach leads to a substantial decrease in the error rate in comparison with other non-learning methods.

## 1. INTRODUCTION

Nowadays, the role of computers has significantly increased in our daily lives. As a result, most of the computerized activities are stored as some sort of data such as text, photos, videos, audio files and more. Hence, it is not surprising that searching for data and finding specific data in all these massive datasets is not only a challenging task in many fields but also quite often a necessary one (Rodríguez et al. 2015).

One of these challenges stems from the Content-Based Image Retrieval (CBIR) which is considered as an important task in "biomedicine, military, commerce, education, and Web image classification and searching. In the biomedical domain, CBIR can be used in patient digital libraries, clinical diagnosis, searching of 2-D electrophoresis gels, and pathology slides" (Wang 2001). CBIR is primarily concerned with searching for and delivering similar images provided a query (input) image is given by a user.

One of the practical aspects of CBIR in medical imaging is to assist clinicians for diagnostic purposes by enabling them to compare the case they are examining with previous (known) cases. It is established pratcice that most hospitals do store their patient data for a long periode of time; generally images are stored in PACS (picture archiving and communication system) and related documents such as biopsy and treatment reports are stored in RIS (radiology information system). Let us assume that a diagnostic case is being inspected, by using a reliable CIBR system clinicians can benefit from analogous cases buried among millions of images, and hence achieve higher diagnostic accuracy based on comparative discrimination with previous (known) cases (Kumar et al. 2013). A large number of CBIR methods exist in the literature. Feature extraction, learning, dictionary approaches, and binary descriptions are among most commonly used techniques to search for similar images.

In this paper, we propose compact features to facilitate fast image retrieval. Our method (Single Projection Radon Barcode, SP-RBC) benefit from the information inherent in single Radon projections, based on the recently introduced Radon Barcodes (RBC) that capture image information in short binary vectors, or barcodes. We will report both cases where we use the actual values of Radon projections as well as their binary encodings known as Radon barcodes (RBC).

The rest of this paper is organized as follows: Section 2 provides a brief review of related works and Radon transform. Section 3 will describe our proposed method. The results of our experiments and comparisons are reported in Section 4. The last section provides some conclusions and suggestions for the feature works.

## 2. RELATED WORK

Content-based image retrieval (CBIR) techniques automatically search for similar images in a database by using visual features extracted to represent its content, and not by a text description. In many cases, textual descriptions may not be available, and in many other cases, it is extremely difficult, if not impossible, to describe the image content (e.g., the shape of an irregular tumor in a breast ultrasound scan) adequately in all necessary details.

The major challenge for CBIR is to extract the relevant image features based on relevant feature similarity criteria and to organize the extracted features into some sort of compact embeddings or representations for fast retrieval from big databases (Kumar et al. 2013). The features or descriptors that represent the properties/content of the images are often used in CBIR systems. The choice of features or descriptors should minimize the "semantic gap" between the extracted image features on one side and the human's interpretation of the image content on the other side.

Early CBIR systems often used the image features/descriptor, such as histogram, shape and texture descriptors (Gevers & Stokman 2004; Lee et al. 2003; Saha et al. 2004). Gevers and Stokman proposed an object recognition method based on the histograms derived from photometric color invariants, which outperformed the traditional color histogram scheme but was very sensible to the noise (Gevers & Stokman, 2004). The edge histogram which contains the general shape information and the moment that describes the image pixel intensities were also used in early CBIR systems (Shim et al. 2002; Zhu & Schaefer 2004).

The advanced features such as Scale Invariant Feature Transform (SIFT) (Lowe 2004) and Speeded Up Robust Features (SURF) (Bay et al. 2008) are employed in CBIR systems to retrieve similar images from different point of views and transformations (Do et al. 2010; Velmurugan & Baboo 2011). As many of these smart features are invariant to scale and rotation, they are more robust than typical image transforms. However, the features are typically large and inefficient to conduct matching in big image data.

Ledwich et al. used the structure of typical indoor environments to reduce the need for rotational invariance of the features, which has a minimal effect on retrieval rate and significant improvement in efficiency (Ledwich & Williams 2004). Velmurugan and Babbo used the KD-tree with the Best Bin First (BBF) (Beis & Lowe 1997) indexing method to accelerate the similarity match of the SURF and color moments combined features (Velmurugan & Baboo 2011).

With dramatic growth of image data in recent years, one of the current trends in CBIR is to use binary features such as Local Binary Patterns (LBP) (Ojala et al. 2002), Binary Robust Invariant Scalable Keypoints (BRISK) (Krig 2012), Binary Robust Independent Elementary Features (BRIEF) (Calonder et al. 2010), and Radon Barcodes (Tizhoosh 2015). Comparing to non-binary features, the distance computation between binary strings is much faster for retrieving tasks. Bankar et al. proposed a CBIR system based on LBP variance which characterized the local contrast information into the one-dimensional LBP histogram (Bankar et al. 2014). Subrahmanyam et al. extended the LBP, which took advantage of the magnitude of the local difference between the center pixel and its neighbors which were able to extract the edge information in the image, and proposed the local maximum edge binary patterns (LMEBP) descriptor (Subrahmanyam et al. 2012).

Other methods such as Deep Neural Networks (DNNs), Convolutional Neural Networks (CNNs), and Bag of Words (BoW) have been recently developed for CBIR tasks. Learning from massive annotated data in the deep learning networks, CNN features would carry high-level and riche semantic information (Yan et al. 2016), which had been proved to be successful in achieving state-of-the-art performance (Babenko et al. 2014; Wan et al. 2014). Avni et al. made use of SIFT descriptors to build the feature dictionary with the bag of visual words approach, performing outstandingly for the x-ray image retrieval task in IRMA dataset (Avni et al. 2011). The best results so far have been reported by combining LBP and saliency detection (Camlica et al. 2015).

However, most of CBIR methods face high computational expenses and require considerable resources during the learning phase. Besides, their implementation requires sophisticated and complicated design and data structures (Krig 2012).

In the medical field, CBIR systems can assist clinicians to make more reliable clinical decisions by retrieving similar (proven) cases from the past stored in their archives. But for the most of the afore-mentioned methods, they usually need to undergo param-eter tuning to be applicable to medical image processing (Huang et al. 2010). Not only because most of the medical images have a known direction or known scale (which means the global search would not face related challenges), but also because at least for the case of local descriptors in medical images, experiments show they are generally not able to achieve good results (Avni et al. 2011). It would be acceptable if we use very simple content-based retrieval methods, rather than learning-based complicated methods in order to save time for learning and parameter tuning, even if we cannot reach top accuracies, but can provide comparable results.

As Radon transform can convert global detection problem in the image domain into local peak detection problem in the parameter domain (Aundal & Aasted 1996), it is widely applied in the medical field, especially in Computed Tomography. Radon barcodes, RBC, are binary codes generated by Radon Transform with projection binarization, proposed for medical image retrieval system. RBCs have achieved comparable results with many other methods from literature (Tizhoosh 2015) and are easy to implement and efficient in matching and retrieving (via Hamming distance) with low requirements for memory and storage.

In this paper, a content-based image retrieval approach for big datasets is proposed. The multiple Radon projections with selected angles are first extracted (considered as global features) for each image. Each projection is used separately to search for similar images from the big dataset. The more precise local features such as LBP and shifted Radon projections are then employed to refine the results. We use the IRMA dataset of 14,400 x-ray images to validate the approach.

## 2.1 Radon Transform

Radon introduced an integral transformation, which calculates the sum of the values of an image along parallel lines for various angles (Radon 1917). One key factor of Radon transform is its ability to reconstruct the main image from its transform. The Radon transform has been applied in medical imaging, e.g., in computer tomography, for image reconstruction (Gu & Sacchi 2009). The Radon transform is generally given as for each given $\theta$ as follows:

$$R(\rho,\theta)=\int_{-\infty}^{+\infty}\int_{-\infty}^{+\infty}f(x,y)\delta(x\cos\theta+y\sin\theta-\rho)dxdy \quad (1)$$

In this equation, $f(x,y)$ refers to grey-level intensities of image $f$ at position $(x,y)$, and $\delta(.)$ is a Dirac delta operator. Figure 1 depicts an example for a small matrix and three sample Radon projections which can be used as image features and have been applied in many fields of computer vision (Aundal & Aasted 1996). It also has been applied in form of Radon Barcodes for medical CBIR systems in large datasets (Tizhoosh 2015).

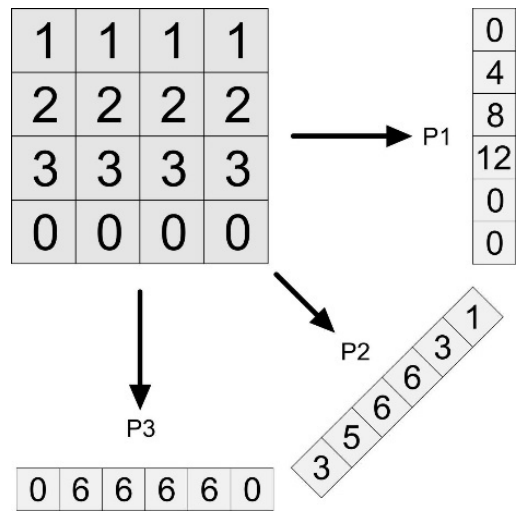

Figure 1. Three Radon projections in 0, 45 and 90-degree directions. Zero padding is applied to create same-length vectors.

## 3. PROPOSED METHOD

In this section, we introduce the idea to use single Radon projections for CBIR in its both forms, namely real-valued (single projection Radon, SP-R) and binary (single projection Radon barcode, SP-RBC) implementations. First, we describe the automated preprocessing steps. In the next step, we explain how we use Radon projections to reach top similarity for each projection separately. We then introduce the exploitation method, which is used to find the most similar images in the pre-selected set (Selection Pool). Finally, we apply a binarization method to create SP-RBC. For all our experiments, we use the IRMA dataset (Tommasi et al. 2009).

### 3.1 Pre-Processing Images

Data test images we use is quite challenging. For instance, the imbalance in IRMA image dataset has been noted as one of the most challenging aspects of this dataset (Avni et al. 2011). There is major variability in IRMA images, not only in the term of a sample density in each category but also with respect to image size, brightness, scale of body objects as well as unrelated burned-in landmarks (Figure 2). To reduce the effects of these problems, we proposed a pre-processing chain composed of three stages:

1) resizing images to zero-padded squared images to avoid distortion due to necessary under-sampling (most CBIR methods do require under-sampling of images to reduce computational burden),

2) removing non-related parts such as burned-in landmarks (e.g., letters) due to the digitization of x-ray films, and

3) circular image margin suppression (super-imposing a circle on the image to eliminate the image margin from the processing).

Figure 3 shows the output of our pre-processing steps when applied on images from Figure 2.

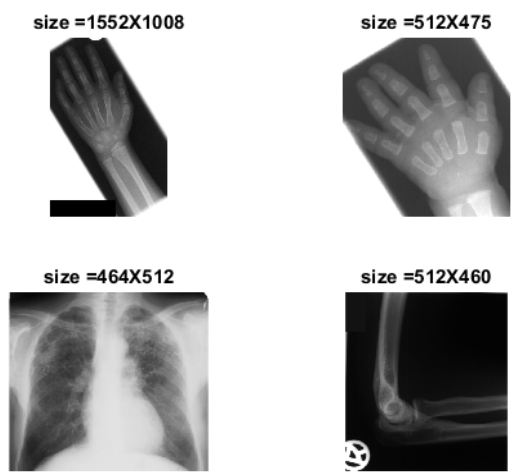

Figure 2. Four samples from IRMA training set to illustrate the variability of X-ray image data.

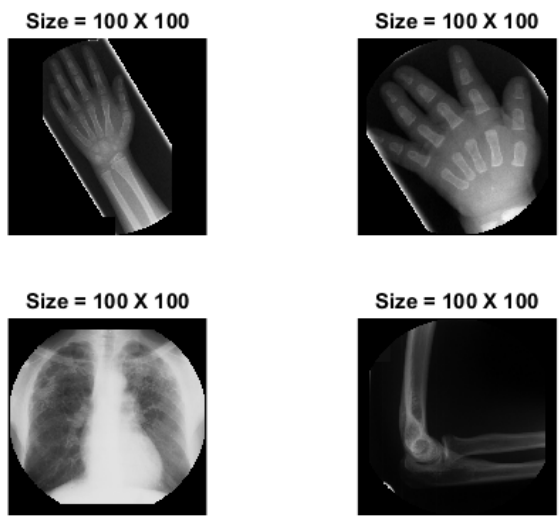

Figure 3. The pre-processing effect for same samples from Figure 2 to create "normalized" inputs.

Figure 4 shows the difference between our squaring method and simple image resize operations. This operation preserves the original scale of image whiteout distortion due to the resize operation which shows significant improvements in the results (see Table 4 in the result section) in comparison with previous works.

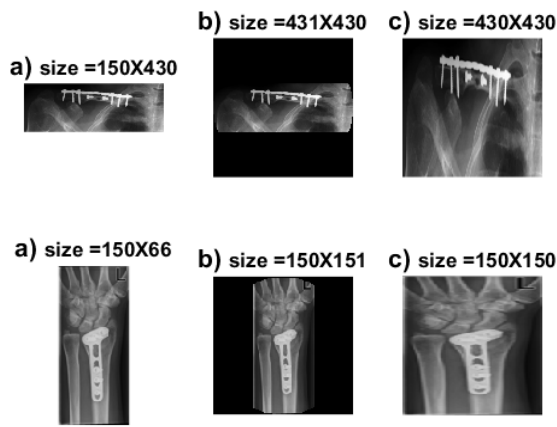

Figure 4. a) rectangular images, b) resized by our method (squared with zero padding), and c) resized conventionally.

For the second pre-processing stage, we consider the nature of summation operator in Radon vectors which is clearly sensitive to bright parts of image. As shown in Figure 2, most of the unrelated shapes (letters, signs, landmarks etc.) in IRMA dataset are depicted in (near-)white. To decrease their effect on Radon projections, we remove these white parts by a simple binary operation. Although this may sometimes lead to eliminating some relevant characteristics of some images. In general, however, our results have shown that this stage of pre-processing helps to deliver better results (see Table 4 in result section). In this stage, all images are binarize by a threshold at 98% of maximum image intensity to locate (near-)white image regions (Figure 5). Subsequently, these parts are replaced (filled) with the median intensity value of their neighbors (we access the neighbors by using some morphological operators). Also, since artificial marks/signs appear mostly close to the image border, replacing (near-)white pixels is strictly restricted to the image margins.

Radon projections change their length based on projection angel. All Radon operators consider the maximum length $\sqrt{2}N$ (where $N$ is the largest of image width/height) and use zero-padding for this purpose. Based on (Tsai & Chiang 2010), we discard all pixels out of circumference area from the image center within a diameter of $N/2$.

Figure 3 shows sample images with superimposed circles. This process helps to achieve shorter vectors;

the assumption, of course, is that image margins do not offer much diagnostic information, hence we can eliminate them. Our experiments confirm the assumption for x-ray images in IRMA dataset. We also can perform this step by selecting just $N$ center elements out of $\sqrt{2}N$ elements in all Radon projections.

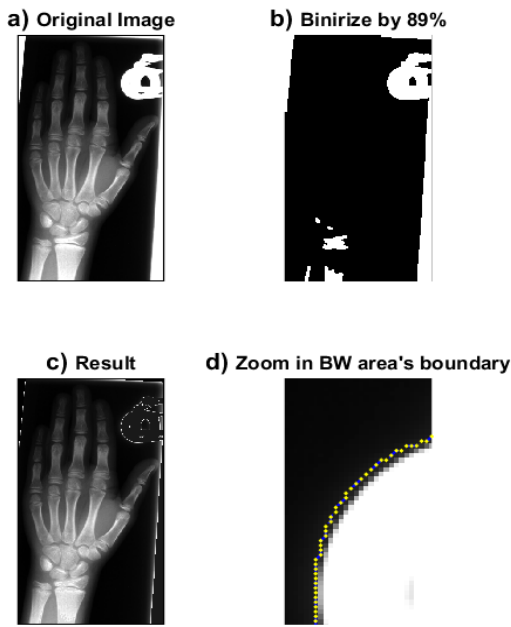

Figure 5. Removing irrelevant (near-)white landmarks: a) original image, b) binary image with 98% thresholding, c) the result after removing the irrelevant pixels by replacing them with the median value of their neighborhood, d) magnification of upper region of the image to show the boundaries which are used for median calculation.

## 3.2 Radon Projections

Using several Radon projections simultaneously has been used successfully in many CBIRs woks (Zhu & Tizhoosh 2016; Tizhoosh 2015; Liu et al. 2016). It may be seen as obvious that using more projections in all possible directions is associated with better results and naturally a more time-consuming search. In our experiments, we examine a new approach by using only one projection for each retrieval attempt. Obviously, we did not anticipate very promising retrieval results for just one projection. However, there was an interesting point in the results. If we, for the sake of analysis, select the minimum IRMA error for each projection, then the total IRMA error decreases dramatically (much better than concatenating them into one vector). It means that although individual projection results are slightly better than of random results (the error of each projection results is varying between 570 to 640, and random search error is around 900, see Table 3). This observation confirms that separated projection results may have little overlap with each other. As a result, while one projection (for example the projection at zero degree) fails to find the most similar image, other projection (say the 45-degree direction) might be able to find it. By choosing a certain number of "best matches" (say the top three matches) for each Radon projection, we can create a "Selection Pool" (see Figure 6).

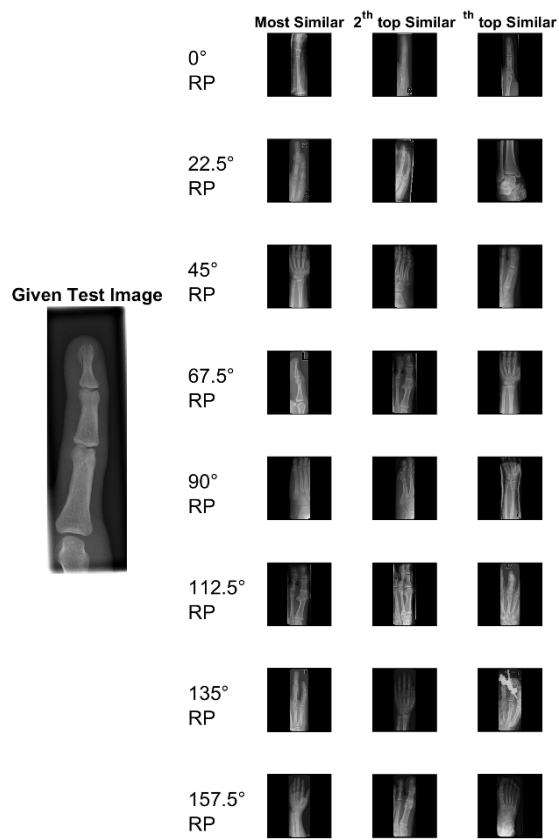

Figure 6. A sample of Selection Pool for a test image and top three images for each Radon projection. The final result(s) can be extracted from the Selection Pool after more computationally demanding search methods are applied.

The surprising point is that the minimum error rate for each test image in the Selection Pool considerably decreases (e.g., error=196 for 8 projections). Figure 7 motivates our idea of using single projections separately, where the error of the first hit (image with the highest similarity to the query image) for 8 projections are depicted. The right side plot in Figure 7 is the magnification of the errors of one specific test image (the image #619 in IRMA dataset). As shown, each projection can be used to retrieve an image as the first hit with different IRMA error. If we record the lowest error for each projection, we can reach a total error of 196 which, looking at the reported numbers in literature, is quite low. For 10 top images for each projection, the error even decreases to 65, an error level not yet reached by any method in literature. It means if we had an algorithm to identify the best Radon projection for each image, we could achieve the outstanding results. However, reliable learning methods to select the best projection is apparently quite challenging and subject to future works. Hence, we attempt to exploit the discrimination power of single projections in a Selection Pool via an exploitative approach.

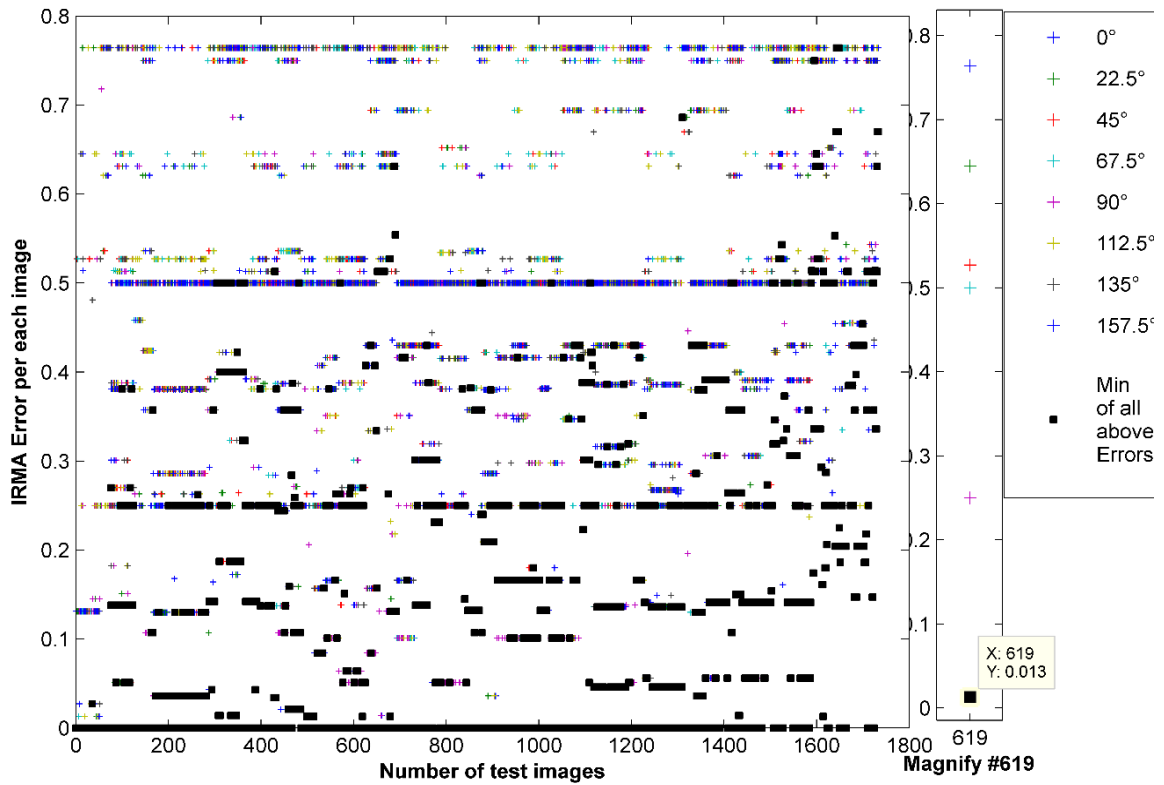

Figure 7. The IRMA error of 1733 images for 8 single equidistant Radon projections. Black squares are the minimum error for each test image. The sum of black squares amounts to 196. The right section magnifies an arbitrary test image results (#619) for better visualization.

## 3.3 Exploitation Search

After we have a small group of candidate images (Selection Pool), retrieved from thousands of images in training set, we can now apply a refined and more exploitative search to choose the best image from within the Selection Pool. As mentioned before, if we manually pick the best image from the top ones, we can achieve the best scores. In this section, two methods are combined to search the Selection Pool; Shifted Radon and LBP.

Shifted Radon is proposed to eliminate the effect of translation in images or image regions. Using eight equi-distanced projections and shifting each Radon projection in the test image to align with its counterpart projection in the Selection Pool makes the algorithm robust against translation (Figure 8 shows this process for two sample images). We use the smallest distance between the two projections using cross-correlation (shifting by ± 10% the length of the projection). By looking at cross-correlation between each pair of Radon projections of two images, all eight minimum distances (we use eight projections) are summed up. After calculating the distance for all images in the Selection Pool, we normalize this error in [0,1].

LBP (local binary patterns) have been used in many CBIRs based on thier power and speed (Avni et al. 2011)(Nanni et al. 2010). In our method, LBP and Shifted Radon are used together to improve the results. The LBP error rate is calculated by the normalized sum of their absolute values. Final decision making is done by using the smallest value of the sum of these two error vectors. Since both vectors (Shifted Radon and LBP) are normalized between zero and one, they are comparable to each other.

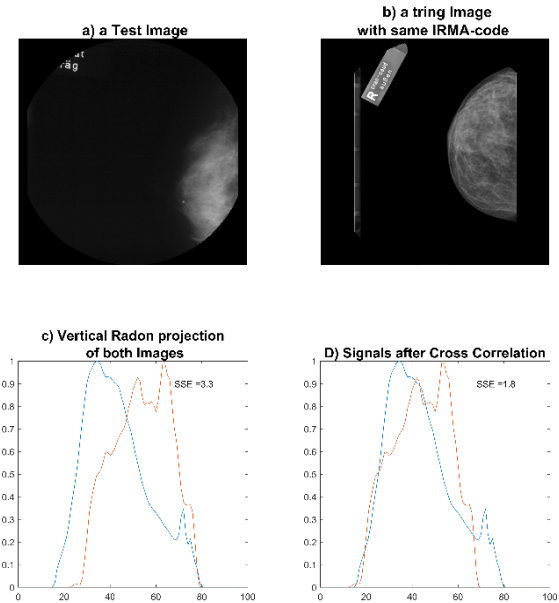

Figure 8. Two sample images with vertical subject shift and their zero-degree Radon projections before and after cross-correlation-based shifting.

## 3.4 Binarization of Projections

Computational complexity is one of the critical features of any retrieval method. Our approach is very light in comparison with most reported descriptors. We just search the dataset using eight vectors as short as the length of the resized image (under-sampled to 64x64 pixels in our experiments). Additionally, to further shorten the search time, we have created the binary version of projections to test SP-RBC which employs two methods to create binary codes from Radon projections. The MinMax and Median methods have already been applied to threshold Radon projections (Tizhoosh 2015, Tizhoosh et al. 2016).

The Median method uses the median value of each projection as a threshold, all elements below the threshold are set to zero (Tizhoosh 2015). On the other hand, the MinMax method sets zeros and ones depending on the locations of minimums and maximums of each projection (Tizhoosh et al. 2016).

## 4. EXPERIMENTS AND RESULTS

This section introduces the data set used in our research. We then detail our results and compare our method with other methods.

### 4.1 Image Dataset: IRMA

The Image Retrieval in Medical Application, short IRMA, is a challenging dataset, which has composed of 12,677 images for training and 1,733 images for evaluating any proposed retrieval method. Every IRMA image is associated with a 13-digit code, and each code is divided into four parts:

IRMA Code: TTTT-DDD-AAA-BBB

The first four digits describe the imaging modality, the next three represent the body orientation in the image, the next four describe the body region and finally, the last four indicate the biological system examined. IRMA creators have also introduced a system for measuring the error between two IRMA codes and return an error number between zero and one (Tommasi et al. 2009). So, retrieval algorithms should return the most similar image for all 1733 test images using the training dataset which supposed to contain similar cases for every query image although due to imbalance the easiness of finding similar cases strongly varies.

### 4.2 Results

In this section, we first compare our best-achieved result with the results of other methods. After that, we discuss the impact of each part of implementation on achieved results. Finally, we show the results for each Radon projection separately.

For comparing with other methods, we consider approaches in two different groups, non-learning methods and learning-based methods. The results for $RBC_4$, $RBC_8$, and LBP are used from literature (Tizhoosh 2015). However, since there is a different way for error calculation in IRMA database than in some papers, we recalculate the error for all mentioned methods based on (Tommasi et al. 2009) to have consistent and fair comparisons.

In non-learning comparison, we consider L as the length of descriptors, T as the type of searching (B for Binary search and F for Floating point search), $E_{Total}$ as total IRMA error, and $N_0$ indicates the percentage of zero-error cases (in case consider the classification problem).

Table 1: IRMA error for non-learning methods (L=length, T=type [B=binary; F=float], E=error, $N_0$=percentage of retrieved cases with zero error).

| Method | L | T | $E_{Total}$ | $N_0$ |
|---|---|---|---|---|
| SP-R | 8×64 | F | 311.80 | 45.76% |
| SP-RBC$_{Min-Max}$ | 8×64 | B | 356.57 | 42.30% |
| SP-RBC$_{Median}$ | 8×64 | B | 419.86 | 34.16% |
| LBP | 1×135 | F | 365.23 | 38.26% |
| RBC$_8$ | 8×64 | B | 580.68 | 25.39% |
| RBC$_{16}$ | 16×64 | B | 564.54 | 23.54% |

Table 1 shows that the proposed method SP-RBC does loose some information in exchange for some increase in speed compared to SP-R (single projection Radon) which simply uses the floating-point projections values without thresholding. But it can be observed that the MinMax method is significantly better than Median thresholding.

We also compared our results with the most successful methods, which are applied on IRMA dataset. Because all of them use some notion of learning, they may only use labeled IRMA test images (94 images in IRMA 2009 are not labelled; some works just ignore them). Hence, their error might increase around 5-6%.

In the next part of this section, we have analyzed the parameter tuning and the details of our observations. Firstly, we discuss the size of the Selection Pool, which can affect the error rate significantly. Figure 9 reflects the relationship between the error rate and the number of top images per projection. As it can be seen, the error rate has dropped substantially between the first hit and the top five, while the decreasing rate tends to remain constant after number of top choices reaches 10 for SP-R. The error rate seems to continue to improve for SP-RBC beyond considering more than 10 top choices. The best answer is reached by looking at top 14 images per projection. It means we search among 112 images in the Selection Pool (we use 8 separate projections).

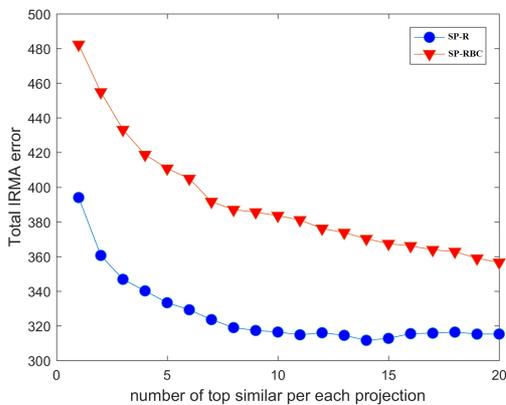

Figure 9. Total IRMA error based on number of top similar images for each projection.

Table 2: Best learning-based methods. All result marked with * are reported in (Tommasi et al. 2009)

| Learning method | $E_{Total}$ |
|---|---|
| (Camlica et al. 2015) | 146.55 |
| TAUbiomed* | 169.50 |
| diap* | 178.93 |
| VPA* | 242.46 |
| FEITIJS* | 261.16 |
| **SP-R** | 311.8 |
| MedGIFT* | 317.53 |
| VPA* | 320.61 |
| **SP-RBC** | 356.57 |
| IRMA* | 359.29 |
| MedGIFT* | 420.91 |

Table 2 shows the results compared to learning-based methods. The best results reported by Camlica et al. is based on extensive saliency detection. The second-best solution uses a dictionary approach accompanied by PCA application.

The results for each separate Radon projection, as well as their binary version are provided in Table 3. Each projection has a relatively high error.

Table 3: Results for SP-R and SP-RBC.

| Learner | SP-R $E_{Total}$ | SP-RBC $E_{Total}$ |
|---|---|---|
| 0-degree Radon | 567.67 | 644.52 |
| 22.5degree Radon | 598.7 | 687.71 |
| 45-degree Radon | 613.46 | 700.15 |
| 67.5-degree Radon | 642.25 | 719.29 |
| 90-degree Radon | 561.38 | 649.21 |
| 112.5-degree Radon | 629.77 | 710.43 |
| 135-degree Radon | 618.41 | 721.03 |
| 157.5-degree Radon | 575.43 | 676.69 |

In Table 4, we share the results of concatenated Radon projections as one vector. We also provide information about the impact of preprocessing steps. Normalization in general improves the results about 5%. Table 4 shows that there is some improvement (approx. 10%) in white part removal and zero-padded square resizing.

Table 4: Eight concatenated Radon projections.

| Type of preprocessing | $E_{Total}$ |
|---|---|
| Radon whiteout preprocessing | 439.93 |
| Normal Radon | 420.82 |
| Normal Radon + resize method | 389.10 |
| Normal Radon + resize method +criclize | 384.68 |
| Normal Radon + removing white spots | 385.75 |
| All preprocessing steps | 383.41 |

## 5. SUMMARY

In this paper, we proposed the idea of using single Radon projections for medical image retrieval in large archives. This can be considered an improvement of previous works (Tizhoosh 2015) which introduced the idea of Radon Barcodes by binarization of a selected number of Radon projections.

In our method, a Selection Pool can be assembled when multiple single Radon projections are applied separately to retrieve many images from the database. The single projection Radon Barcodes (SP-RBCs) may lose some information due to the thresholding process but they are compact and fast for retrieval in big image data. Subsequently, more time consuming local search can be performed on the Selection Pool to retrieve the most similar cases. In this paper, we employed LBP and Shifted Radon but many other alternatives could be investigated.

In our future work, we will focus on improving the exploitative search in the Selection Pool and learning method to find the best projection to search data set by just one projection.